\documentstyle[lrec2000]{article}

\title{A Novelty-based Evaluation Method for Information Retrieval}

\name{Atsushi Fujii, Tetsuya Ishikawa}

\address{University of Library and Information Science \\
1-2 Kasuga Tsukuba 305-8550, JAPAN \\
{\{fujii, ishikawa\}@ulis.ac.jp}}

\abstract{In information retrieval research, precision and recall have
long been used to evaluate IR systems. However, given that a number of
retrieval systems resembling one another are already available to the
public, it is valuable to retrieve novel relevant documents, i.e.,
documents that cannot be retrieved by those existing systems. In view
of this problem, we propose an evaluation method that favors systems
retrieving as many novel documents as possible. We also used our
method to evaluate systems that participated in the IREX workshop.}

\newcommand{\eq}[1]{(\ref{#1})}

\begin{document}

\maketitleabstract

\section{Introduction}
\label{sec:introduction}

In information retrieval (IR) research, the notion of precision and
recall have commonly been used to evaluate the empirical performance
of systems~\cite{keen:ipm-92,salton:ipm-92}. Precision is the ratio of
the number of relevant documents retrieved by a system under
evaluation, compared to the total number of documents retrieved by the
system. On the other hand, recall is the ratio of the number of
relevant documents retrieved by the system, compared to the total
relevant documents in a given benchmark test collection.

In other words, the precision/recall-based evaluation method regards
all the relevant documents as equally important or informative for the
user, and thus highly values systems that retrieve as many relevant
documents as possible, with little noise.

However, in the real world, where a number of IR systems are
available, for example, on the World Wide Web, it is often the case
that the user has already read some of relevant documents using other
systems. Thus, systems that always retrieve relevant documents similar
to those retrieved by ubiquitous systems have little practical
utility. In addition, meta search systems, which integrate document
sets retrieved by more than one system, are less effective, in the
case where individual systems retrieve similar documents.

In view of these problems, our proposed IR evaluation method favors
systems that retrieve more {\em novel\/} documents, that is, relevant
documents which cannot be retrieved by other existing systems.

From a different perspective, our evaluation method is also effective
in producing test collections. The pooling
method~\cite{voorhees:sigir-98}, which has commonly been used to
produce test collections, requires a variety of participating systems.
However, in the case where most participating systems adopt similar
techniques, it is not feasible to collect a sufficient ``pool'' (i.e.,
a set of candidates for relevant documents).  Our evaluation method is
expected to promote a development of IR systems with various concepts,
and therefore resolve the above problem.

Section~\ref{sec:measure} formalizes the evaluation measure based on
the novelty of documents, and Section~\ref{sec:case_study} applies
this measure to evaluate IR systems that participated in the IREX
workshop~\cite{sekine:irex-99}.

\section{Formalizing the Measure}
\label{sec:measure}

Instead of the notion of precision and recall, we propose as a new
evaluation measure the utility of system $x$ with respect to relevant
document $d$, \mbox{$U_{d}(x)$}. This measure denotes the extent to
which $x$ contributes to providing the user with $d$, for a given
query.  Note that in this paper, $d$ generally refers to a {\em
relevant\/} document.

From an information theoretical point of view, we calculate
\mbox{$U_{d}(x)$} as the ratio of the probability that the user reads
document $d$ by using system $x$, \mbox{$P(D=d|S=x)$}, compared to the
probability that the user reads $d$ by using another system (i.e.,
even without using $x$), \mbox{$P(D=d)$}, as shown in
Equation~\eq{eq:udx}.
\begin{equation}
  \label{eq:udx}
  U_{d}(x) = \log\frac{\textstyle P(D=d|S=x)}{\textstyle P(D=d)}
\end{equation}
In the case where system $x$ adopts a ubiquitous retrieval technique,
the value of \mbox{$P(D=d|S=x)$} becomes similar to that of
\mbox{$P(D=d)$}, and thus the utility of $x$ becomes small.  On the
other hand, the utility of $x$ becomes greater as the number of {\em
novel \/} relevant documents provided by $x$ increases.

We then calculate the {\em total\/} utility of $x$, $U(x)$, by summing
up $U_{d}(x)$'s of all the relevant documents for the query, as shown
in Equation~\eq{eq:ux}.
\begin{equation}
  \label{eq:ux}
  U(x) = \sum_{d} U_{d}(x)
\end{equation}
To sum up, our evaluation method favors systems with greater
\mbox{$U(x)$}.

In Equation~\eq{eq:udx}, \mbox{$P(D=d)$} is the summation of
\mbox{$P(D=d|S=y)$}'s for existing systems, averaged by the
probability that the user utilizes system $y$, \mbox{$P(S=y)$}.  Thus,
given a set of existing system excluding $x$, $E$, we calculate
\mbox{$P(D=d)$} as in Equation~\eq{eq:pd}.
\begin{eqnarray}
  \label{eq:pd}
  \begin{array}{lll}
    P(D=d) & = & {\displaystyle \sum_{y\in E}P(D=d|S=y)\cdot P(S=y)} \\
    \noalign{\vskip 2ex}
    & \approx & {\displaystyle \sum_{y\in
    E}P(D=d|S=y)\cdot\frac{\textstyle 1}{\textstyle |E|}}
  \end{array}
\end{eqnarray}
Here, note that we assume uniformity with respect to \mbox{$P(S=y)$}.

Finally, the crucial content is the way to estimate
\mbox{$P(D=d|S=x)$}, i.e., the probability that the user reads
document $d$ by using system $x$. It can safely be assumed that the
user always reads the top document, $d_1$, and thus $P(D=d_{1}|S=x)$
always takes 1. However, the probability that the user reads remaining
documents becomes smaller according to their ranking.

Given $N$ documents sorted according to their relevance degree, in
descending order, the user can choose a threshold for the ranking
(i.e., the boundary until which he/she continues to read) out of $N$
choices. Consequently, documents ranked lower than the threshold will
be discarded.

In other words, we can calculate \mbox{$P(D=d|S=x)$} as the
probability that the user chooses a threshold equal to or greater than
the ranking of $d$, as in Equation~\eq{eq:pdx}.
\begin{equation}
  \label{eq:pdx}
  \begin{array}{lll}
    P(D=d|S=x) & = & {\displaystyle \sum_{i = r_{x,d}}^{N}
    \frac{\textstyle 1}{\textstyle N}} \\
    \noalign{\vskip 2ex}
    & = & \frac{\textstyle N - r_{x,d} + 1}{\textstyle N}
  \end{array}
\end{equation}
Here, $r_{x,d}$ is the ranking of document $d$ determined by system
$x$.

\section{A Case Study using the IREX Collection}
\label{sec:case_study}

Our concern in this section is to investigate the characteristic of
our evaluation method. For this purpose, we targeted IR systems
participated in the IREX workshop~\cite{sekine:irex-99}, and compared
the result obtained based on our newly proposed evaluation method,
with that based on the precision/recall. We also investigated reasons
behind the difference between those two results, if any.

\subsection{Overview of the IREX Collection}
\label{subsec:irex}

The IREX collection was produced through the IREX
workshop~\cite{sekine:irex-99}, which consists of TREC-style IR and
MUC-style named entity (NE) tasks for Japanese.\footnote{{\tt
http://cs.nyu.edu/cs/projects/proteus/irex/\\index-e.html}} Hereafter,
the IREX collection/workshop refers solely to that related to the IR
task.

The IREX collection consists of 30 queries, 211,853 articles collected
from two years worth of ``Mainichi Shimbun'' newspaper
articles~\cite{mainichi:94-95},\footnote{Practically speaking, the
IREX collection provides only article IDs, which corresponds to
articles in Mainichi Shimbun newspaper CD-ROM'94-'95. Participants
must get a copy of the CD-ROMs themselves.} relevance assessment for
each query, retrieval results of 22 participating systems, and
technical details of each system.

Each query consists of the ID, description and narrative.  While
descriptions are usually phrases to briefly express the topic,
narratives consist of several sentences and synonyms associated with
the topic. Figure~\ref{fig:query} shows an example query in the SGML
form (translated into English by one of the organizers of the IREX
workshop).

\begin{figure}[htbp]
  \begin{center}
    \leavevmode
    \small
    \begin{quote}
      \tt
      <TOPIC> \\
      <TOPIC-ID>1001</TOPIC-ID> \\
      <DESCRIPTION>Corporate merging</DESCRIPTION> \\
      <NARRATIVE>The article describes a corporate merging and in the
      article, the name of companies have to be identifiable. Information
      including the field and the purpose of the merging have to be
      identifiable. Corporate merging includes corporate acquisition,
      corporate unifications and corporate buying.</NARRATIVE> \\
      </TOPIC>
    \end{quote}
    \caption{An example query in the IREX collection.}
    \label{fig:query}
  \end{center}
\end{figure}

Relevance assessment was performed based on the pooling
method~\cite{voorhees:sigir-98}. That is, candidates for relevant
documents were first pooled using the 22 participating systems.
Thereafter, for each candidate document, human experts assigned one of
three ranks of relevance, i.e., ``relevant'', ``partially relevant''
and ``irrelevant''.  The average number of documents pooled for each
query is 2,105, among which the number of relevant and partially
relevant documents are 68 and 116, respectively.

Each retrieval result consists of the top 300 articles submitted in
the same form as used in the TREC.\footnote{{\tt
http://trec.nist.gov/pubs.html}} For each of the 22 results, the TREC
evaluation software was used to investigate the performance (e.g.,
non-interpolated average precision).  Figure~\ref{fig:trec} shows a
fragment of the retrieval result obtained with one of the
participating systems, which consists of the query ID, dummy field,
article ID, ranking of the article, relevance degree computed by the
system, and system ID.

\begin{figure}[htbp]
  \begin{center}
    \leavevmode
    \small
    \tt
    \begin{tabular}{llllll}
      1007 & 0 & 940228106 & 1 & 0.306856 & 1106 \\
      1007 & 0 & 940110130 & 2 & 0.246505 & 1106 \\
      1007 & 0 & 950106119 & 3 & 0.237173 & 1106 \\
      1007 & 0 & 940131126 & 4 & 0.236115 & 1106 \\
      1007 & 0 & 940614009 & 5 & 0.223313 & 1106 \\
      1007 & 0 & 940614002 & 6 & 0.222998 & 1106 \\
      1007 & 0 & 941107114 & 7 & 0.217324 & 1106 \\
      1007 & 0 & 940428222 & 8 & 0.215979 & 1106
    \end{tabular}
    \caption{A fragment of the retrieval result of system ``1106''.}
    \label{fig:trec}
  \end{center}
\end{figure}

\begin{table*}[htbp]
  \tabcolsep=3pt
  \begin{center}
    \leavevmode
    \small
    \begin{tabular}{ll} \hline\hline
      {\hfill\centering Question\hfill} &
      {\hfill\centering Answers\hfill} \\ \hline
      query information used & only description (8), 
      description+narrative (14) \\
      indexing method & word (9), n-gram (3), word+character (2),
      character (1), syntactic phrase (1), \\
      & statistical phrase (1) \\
      proper noun identification & yes (5) \\
      query expansion & local feedback (2), use of a thesaurus (2) \\
      retrieval method & vector space model (13), probabilistic model
      (4), latent semantic indexing (1) \\
      \hline
    \end{tabular}
    \caption{A fragment of the result of the IREX questionnaire.}
    \label{tab:spec}
  \end{center}
\end{table*}

It should be noted that using relevance assessment and retrieval
results for each system, we can easily calculate \mbox{$P(D=d|S=x)$}
in Equation~\eq{eq:pdx}, which is the central issue in estimating our
evaluation measure.

Technical details of participating systems were collected from
questionnaires answered by each participant, where questions ranged
from retrieval algorithms used to execution time. Although several
questions are relatively vague, a number of questions are effective to
characterize each system.

Table~\ref{tab:spec} shows representative questions in terms of
retrieval accuracy. In this table, the number of answers are indicated
in parentheses. However, answers classified as ``no'', ``unknown'' and
``etc.'' are not shown. Roughly speaking, most systems adopted the
word-based indexing and vector space model combined with TF$\cdot$IDF
term weighting.

On the other hand, note that in the IREX workshop, the correspondence
between system IDs and participants is not available to the
public. Additionally, several participants did not have oral
presentations and papers in the proceedings. Consequently, for some
systems it is difficult to obtain sufficient technical details.

For example, although most participants answered ``TF$\cdot$IDF'' for
the question about term weighting method, it is not possible to
identify the exact formula used, out of a number of
variants~\cite{salton:ipm-88,zobel:sigir-forum-98}, for several
systems.

\subsection{Experimentation}
\label{subsec:experiment}

As explained in Section~\ref{subsec:irex}, the 22 IREX participating
systems have already been ranked based on the conventional
precision/recall, using the TREC evaluation software.

Thus, we re-evaluated the 22 systems based on our evaluation method,
and compared results derived from different evaluation methods. To put
it more precisely, we conducted 22 trials in each of which a different
system was under evaluation and the rest were regarded as existing
systems. That is, the former and latter correspond to $x$ and $E$ in
Section~\ref{sec:measure}, respectively.

Note that in this evaluation, we did not regard ``partially relevant''
documents as relevant ones, because interpretation of ``partially
relevant'' is not fully clear to the authors.

Table~\ref{tab:all_A} compares rankings obtained based on
non-interpolated average precision and the utility factor we proposed
in this paper. Table~\ref{tab:qbq_A} compares rankings obtained with
two evaluation methods on a query-by-query basis, where we show solely
the difference of rankings for enhanced readability. Since in the IREX
collection, every query ID consists of four digits stating with
``10'', we simply show the remaining two digits in
Table~\ref{tab:qbq_A}.

\begin{table}[htbp]
  \begin{center}
    \leavevmode
    \small
    \begin{tabular}{cccc} \hline\hline
      System ID &
      {\hfill\centering Avg. Precision\hfill} &
      {\hfill\centering Utility\hfill} &
      {\hfill\centering Difference\hfill} \\ \hline
      1144b & 2 & 1 & +1 \\
      1135a & 3 & 2 & +1 \\
      1144a & 1 & 3 & -2 \\
      1135b & 4 & 4 & 0 \\
      1103b & 5 & 5 & 0 \\
      1106 & 17 & 6 & +11 \\
      1145b & 16 & 7 & +9 \\
      1122b & 7 & 8 & -1 \\
      1103a & 10 & 9 & +1 \\
      1128b & 9 & 10 & -1 \\
      1142 & 6 & 11 & -5 \\
      1122a & 8 & 12 & -4 \\
      1110 & 11 & 13 & -2 \\
      1133a & 19 & 14 & +5 \\
      1133b & 18 & 15 & +3 \\
      1128a & 12 & 16 & -4 \\
      1120 & 14 & 17 & -3 \\
      1145a & 13 & 18 & -5 \\
      1112 & 15 & 19 & -4 \\
      1146 & 20 & 20 & 0 \\
      1132 & 22 & 21 & +1 \\
      1126 & 21 & 22 & -1 \\
      \hline
    \end{tabular}
    \caption{Comparison of rankings obtained based on
    non-interpolated average precision and utility factor.}
    \label{tab:all_A}
  \end{center}
\end{table}

\begin{table*}[htbp]
  \tabcolsep=3pt
  \begin{center}
    \leavevmode
    \scriptsize
    \begin{tabular}{lrrrrrrrrrrrrrrrrrrrrrrrrrrrrrr} \hline\hline
      & \multicolumn{30}{c}{Query ID} \\ \cline{2-31}
      System ID & 07 & 08 & 09 & 10 & 11 & 12 & 13 & 14 & 15 & 16 & 17 &
      18 & 19 & 20 & 21 & 22 & 23 & 24 & 25 & 26 & 27 & 28 & 29 & 30 &
      31 & 32 & 33 & 34 & 35 & 36 \\ \hline
      ~~~1103a & 8 & -7 & 14 & 0 & 8 & 3 & 3 & -14 & 1 & 13 & 5 & -3 & 0
      & -4 & -2 & 3 & -6 & -3 & 6 & 1 & -2 & 13 & 2 & 14 & -3 & -5 &
      -7 & -2 & -3 & 3 \\
      ~~~1103b & -2 & -5 & 6 & 4 & -1 & -3 & -6 & -9 & 4 & -5 & -1 & 1 &
      -3 & -2 & -1 & 8 & 0 & -2 & 1 & -2 & -1 & 7 & 1 & -3 & -5 & -1 &
      -6 & -3 & -2 & 5 \\
      ~~~1106 & 8 & -4 & -9 & -2 & 9 & -2 & 7 & 11 & 5 & -1 & -2 & -4 & 5
      & 4 & 0 & -3 & -3 & 2 & 0 & 0 & -1 & -1 & 1 & 2 & 1 & 2 & 0 & 2
      & 17 & 0 \\
      ~~~1110 & 6 & -1 & -4 & 4 & -1 & 9 & -4 & -10 & -1 & 0 & 4 & -2 &
      -5 & -1 & 0 & 3 & 0 & -2 & -1 & 0 & 0 & 16 & 13 & -1 & -3 & -3 &
      8 & 1 & 3 & -2 \\
      ~~~1112 & -2 & -5 & 0 & 0 & -5 & 3 & -3 & 1 & -11 & 0 & 5 & -5 & 12
      & -2 & -1 & 5 & -3 & -4 & -3 & -1 & -1 & -4 & -6 & -4 & 3 & 1 &
      -4 & -2 & 0 & 0 \\
      ~~~1120 & 1 & -2 & -2 & -1 & 0 & -3 & 4 & -8 & -1 & 0 & 5 & -2 & 7
      & 1 & 0 & 5 & 0 & 2 & 0 & 2 & 0 & -3 & -1 & -1 & 2 & 2 & 6 & 5 &
      -1 & 0 \\
      ~~~1122a & -2 & 2 & -2 & -7 & -5 & 5 & -5 & -11 & -1 & -5 & 1 & 8 &
      -1 & -6 & -2 & -8 & 1 & 1 & 0 & -1 & 4 & -4 & 1 & -1 & -3 & -1 &
      3 & -2 & -3 & -1 \\
      ~~~1122b & -5 & 0 & -8 & 1 & 0 & -8 & 1 & -5 & -9 & -5 & 0 & -2 &
      -3 & -6 & 1 & -4 & 4 & 0 & -2 & 1 & 7 & -3 & -2 & -4 & -4 & 0 &
      6 & 0 & -1 & -2 \\
      ~~~1126 & 0 & 4 & -10 & 0 & 0 & -2 & 0 & 3 & -1 & -1 & -1 & 1 & -1
      & 0 & 0 & 0 & 0 & 0 & 0 & 1 & 1 & 0 & -2 & -3 & 0 & 0 & -3 & -1
      & 0 & 0 \\
      ~~~1128a & -1 & -1 & 4 & -2 & -3 & 0 & 3 & -6 & -8 & -1 & -3 & 4 &
      2 & 9 & 1 & -13 & 0 & 6 & 2 & -1 & 0 & -2 & 1 & 0 & -1 & 1 & 4 &
      -4 & 0 & 4 \\
      ~~~1128b & -2 & 14 & -4 & -4 & -7 & -5 & 11 & 9 & -2 & -2 & -5 & 4
      & -1 & 3 & -2 & -13 & -1 & 1 & 2 & 2 & 0 & 1 & 0 & -5 & 1 & -1 &
      0 & -4 & 0 & -1 \\
      ~~~1132 & 0 & 16 & -9 & 2 & 0 & 0 & 0 & 12 & 21 & 0 & 0 & 10 & 0 &
      8 & 15 & 0 & -4 & 0 & 0 & 0 & 0 & 0 & 2 & 0 & 0 & -1 & 0 & 13 &
      0 & 0 \\
      ~~~1133a & -2 & -2 & -4 & 0 & 3 & 2 & 3 & 15 & 11 & 1 & -5 & -1 & 1
      & 7 & -1 & 3 & 4 & 1 & 4 & 1 & 0 & -2 & -1 & 1 & 4 & 7 & -1 & 0
      & 0 & 1 \\
      ~~~1133b & -3 & -2 & -4 & 2 & 3 & 1 & 11 & 15 & 3 & 0 & -4 & 2 & 0
      & 5 & 1 & 6 & 5 & 0 & 3 & 1 & 0 & -3 & -5 & -1 & 10 & 3 & -2 &
      -2 & 1 & -1 \\
      ~~~1135a & -1 & -2 & 9 & -2 & 4 & -11 & -6 & 4 & 9 & 2 & -6 & -4 &
      -1 & -1 & -1 & -2 & -3 & -1 & -1 & -1 & 0 & -2 & -2 & 0 & 1 & -1
      & -1 & 0 & -1 & -3 \\
      ~~~1135b & 2 & 0 & 6 & -1 & -12 & -13 & -6 & 1 & 2 & 0 & -3 & 1 &
      -5 & -6 & -3 & -1 & -3 & -2 & 0 & -1 & -4 & -7 & -2 & 0 & 0 & -2
      & -1 & -7 & -2 & 0 \\
      ~~~1142 & -4 & -1 & 10 & 0 & -5 & -1 & -7 & -14 & -7 & -3 & -2 & -3
      & -4 & -7 & -5 & -2 & 4 & -3 & -3 & -1 & -2 & -2 & -2 & -5 & 2 &
      -6 & -7 & -6 & -1 & -4 \\
      ~~~1144a & -2 & -1 & -1 & 3 & -1 & 5 & -16 & -9 & -3 & 5 & 1 & -6 &
      -1 & -2 & 0 & 6 & -1 & -2 & -2 & -3 & 0 & 0 & -2 & -1 & 0 & -4 &
      7 & 2 & -1 & -1 \\
      ~~~1144b & -2 & 3 & -1 & 2 & -2 & 5 & -16 & -5 & -2 & 5 & 2 & -5 &
      2 & -2 & 1 & 5 & -3 & 1 & 1 & -1 & 0 & 0 & -5 & -2 & 0 & 1 & 4 &
      2 & -1 & 2 \\
      ~~~1145a & 0 & -4 & -7 & -4 & -5 & -1 & 5 & 11 & -2 & -1 & -1 & -3
      & -1 & -1 & -1 & 1 & 8 & -3 & -5 & 5 & -1 & -4 & 5 & 6 & -2 & 2
      & -4 & -3 & 1 & -3 \\
      ~~~1145b & 3 & -3 & -5 & 5 & 13 & 7 & 12 & 13 & -5 & -1 & -2 & 8 &
      -3 & 4 & 0 & 2 & 1 & 1 & -2 & 0 & -1 & 0 & 5 & 6 & -2 & 7 & 0 &
      13 & -5 & 0 \\
      ~~~1146 & 0 & 1 & 21 & 0 & 7 & 9 & 9 & -4 & -3 & -1 & 12 & 1 & 0 &
      -1 & 0 & -1 & 0 & 7 & 0 & -2 & 1 & 0 & -1 & 2 & -1 & -1 & -2 &
      -2 & -1 & 3 \\
      \hline
    \end{tabular}
    \caption{Query-by-query comparison of rankings obtained based on
    non-interpolated average precision and utility factor.}
    \label{tab:qbq_A}
  \end{center}
\end{table*}

\subsection{Discussion}
\label{subsec:discussion}

Looking at Table~\ref{tab:all_A}, one may notice that rankings of
systems ``1106'', ``1145b'', ``1133a'' and ``1133b'' were
significantly improved within our evaluation method. Thus, we
investigated properties that characterize each of those four systems,
in a comparison with other systems.

First, we found that ``1106'' adopted a relatively simple
implementation, while most systems used more elaborate ones. To put it
more precisely, morphological analysis was performed, and nouns/verbs
were extracted for a word-based indexing. For term weighting, a
TF$\cdot$IDF formula as in Equation~\eq{eq:tf_idf} was used, while
most systems used different methods, such as the logarithmic TF
formulation as in Equation~\eq{eq:log_tf_idf} and one proposed by
Robertson and Walker~\shortcite{robertson:sigir-94}.
\begin{equation}
  \label{eq:tf_idf}
  f_{t,d}\cdot\log\frac{\textstyle N}{\textstyle n_{t}} \\
\end{equation}
\begin{equation}
  \label{eq:log_tf_idf}
  (1 + \log f_{t,d})\cdot\log\frac{\textstyle N}{\textstyle n_{t}}
\end{equation}
Here, $f_{t,d}$ denotes the frequency that term $t$ appears in
document $d$, and $n_{t}$ denotes the number of documents containing
term $t$. $N$ is the total number of documents in the collection.

Second, ``1145b'' conducted a query expansion~\cite{qiu:sigir-93},
while a few systems used query expansion (e.g., one based on a
thesaurus). In addition, a term weighing method based on mutual
information between two terms was introduced. Possible rationales
behind this method include that two terms frequently co-occur are
effective to characterize the domain of documents, and are thus
assigned with greater term weights.

Third, ``1133a'' and ``1133b'' also used domain knowledge for term
weighting. However, unlike the case of ``1145b'', they regarded pages
of news articles as domain. In practice, a greater weight is assigned
to terms whose distribution varies more strongly depending on the
page, because they are expected to characterize the domain. On the
other hand, terms commonly appear in more pages are assigned with a
lesser weight.

To sum up, our novelty-based evaluation revealed the effectiveness of
those properties above, specifically term weighting methods introduced
in ``1145b'', ``1133a'' and ``1133b'', which were overshadowed or
underestimated within the precision/recall-based evaluation.

We devote a little space to consider Table~\ref{tab:qbq_A} for further
investigation. We arbitrarily regarded improvements above seven as
significant, and focused solely on systems with relatively many
significant improvements, that is, ``1103a'' and ``1132''. Although
``1145b'' is associated with the same number of significant
improvements as ``1132'', we previously discussed system ``1145b''
above.

We found that ``1103a'' is one of five systems that conducts a proper
noun identification, and that five of six queries where ``1103a''
achieved significant improvements are directly or indirectly
associated with proper nouns.

Samples of query descriptions directly and indirectly related to
proper nouns include ``1016: Nick Price (a golfer)'' and ``1011:
arrest of suspects of robbery in the {\it Kanto\/} region'',
respectively. Note that in the latter (indirect) case, Japanese
prefectures within the ``{\it Kanto\/}'' region, which are not
explicitly described in the query (e.g., ``{\it Tokyo\/}'' and ``{\it
Kanagawa\/}''), must be identified in news articles.

Finally, ``1132'' is the only system that used Latent Semantic
Indexing (LSI), which is an extension of the vector space model, so as
to retrieve relevant documents including no common terms in a given
query. While as shown in Table~\ref{tab:all_A}, ``1132'' had the
lowest ranking in terms of the average precision, our evaluation
method indicated that in many cases (queries) an LSI-based method is
expected to retrieve relevant documents that other types of methods
fail to retrieve.

\section{Conclusion}
\label{sec:conclusion}

Evaluation methods based on precision and recall have long been used
in information retrieval (IR) research, where systems that retrieve as
many relevant documents as possible are usually highly valued.

However, given the fact that a number of retrieval systems resembling
one another are available to the public (not only in laboratories), it
is valuable to retrieve relevant documents that can never be retrieved
by those existing systems. This notion is also true in various
contexts that require a variety of IR systems, such as meta search
systems and the pooling method in producing IR test collections.

In consideration of these factors, we proposed a new evaluation method
for IR, which favors systems that retrieve more novel documents, i.e.,
relevant documents that many systems fail to retrieve. To realize this
notion, we estimated the utility of a system in question by comparing
the probability that the user reads relevant documents by using the
system, and the probability that the user can read those documents
even without using the system.

We also applied our evaluation method to the 22 systems that
participated in the IREX workshop, and identified several effective
techniques that have been underestimated in the conventional
precision/recall-based evaluation method.

\bibliographystyle{acl}

\section*{Acknowledgments}

The authors would like to thank organizers and participants of the
IREX workshop for their support with the IREX collection.

\end{document}